\documentclass[10pt,twocolumn,letterpaper]{article}

\usepackage{ijcb}
\usepackage{times}
\usepackage{epsfig}
\usepackage{graphicx}
\usepackage{amsmath}
\usepackage{amssymb}
\usepackage{subcaption}
\usepackage{placeins}
\usepackage[ruled,vlined]{algorithm2e}
\usepackage[table,xcdraw]{xcolor}
\usepackage{siunitx} 
\usepackage{multirow}

\newcommand{\imgStub}[3]{\begin{minipage}{#1\textwidth}\begin{center}
       \includegraphics[width=1\textwidth]{#2}\\
#3\end{center}\end{minipage} \hfil}

\ijcbfinalcopy 

\begin{document}

\title{Multi-spectral Facial Landmark Detection}


\author{Jin Keong, Xingbo Dong, Zhe Jin\\
School of Information Technology\\
Monash University Malaysia\\
47500 Subang Jaya, Selangor, Malaysia\\
{\tt\small {\{keong.jin, xingbo.dong, jin.zhe\}@monash.edu}}
\and 
Khawla Mallat, Jean-Luc Dugelay\\
Department of Digital Security \\
EURECOM\\
Sophia-Antipolis, France \\
{\tt\small \{mallat,dugelay\}@eurecom.fr}
}

\maketitle
\thispagestyle{empty}

\begin{abstract}
Thermal face image analysis is favorable for certain circumstances. For example, illumination-sensitive applications, like nighttime surveillance; and privacy-preserving demanded access control. However, the inadequate study on thermal face image analysis calls for attention in responding to the industry requirements. Detecting facial landmark points are important for many face analysis tasks, such as face recognition, 3D face reconstruction, and face expression recognition. In this paper, we propose a robust neural network enabled facial landmark detection, namely Deep Multi-Spectral Learning (DMSL). Briefly, DMSL consists of two sub-models, i.e. face boundary detection, and landmark coordinates detection. Such an architecture demonstrates the capability of detecting the facial landmarks on both visible and thermal images. Particularly, the proposed DMSL model is robust in facial landmark detection where the face is partially occluded, or facing different directions. The experiment conducted on Eurecom's visible and thermal paired database shows the superior performance of DMSL over the state-of-the-art for thermal facial landmark detection. In addition to that, we have annotated a thermal face dataset with their respective facial landmark for the purpose of experimentation.
\end{abstract}

\section{Introduction}

In recent years, infrared spectrum (IR) imagery has been utilized to facilitate the development of computer vision (CV) applications mainly due to its light-insensitive nature. For instance, IR camera can capture images under unfavorable illumination situations, hence it can be used at night or even under completely dark conditions. Most conventional IR imagery today uses an active approach which senses reflected electromagnetic radiation (EMR) within the Near infrared (NIR) or Short-Wave infrared (SWIR) spectrum. This approach relies on the external IR illumination \cite{maldague2002introduction}. As a result, it can be deteriorated by different illumination conditions. Thermal, a.k.a Long-Wave infrared (\SI{7}{\micro\metre} - \SI{14}{\micro\metre} wavelength), is regarded as a promising direction to extend imaging technology \cite{ghiass2014infrared}. Thermal imagery is a passive IR technology that primarily captures the emitted EMR specifically the heat energy from the object, hence thermal spectrum is insensitive to illumination variations. However, it can be affected by ambience temperature.

Bourlai and Cukik \cite{bourlai2012multi} presented some scenarios which require images captured from multiple electromagnetic spectra to perform facial recognition. This suggests that multi-spectral imagery technology can be employed to tackle some CV problems that involve working under a variety of illumination conditions. Chang et al. \cite{chang2008multispectral} showed that the images from multiple spectra can be fused together to perform face recognition under poor illumination conditions. They also presented  a way to fuse visible and thermal face images. There are some scenarios where the visible and thermal imagery can work in conjunction with each other for a more robust performance than a single spectral system. For instance, in real time facial image analysis, we can have a dual camera setup equipped with visible and thermal cameras such as (FLIR Duo Pro R 640), this camera has a thermal sensor as well as visible light sensor designed for camera drone. Depending on the illumination conditions, we can switch in between both configurations, or combine the input to perform a seamless analysis.

One of the main prerequisites of facial image analysis is facial landmark detection as it plays a key role in many tasks, such as automatic face recognition \cite{wiskott1997face} \cite{husken2005strategies}, expression recognition \cite{jung2015joint}, 3D face reconstruction \cite{choi20103d}. Many facial landmark detectors have been designed for face images in the visible spectrum and they are proved to be very reliable. However, these detectors fail to achieve good performance for thermal images \cite{Chu2019}. To achieve multi-spectral analysis, there is a need for an efficient thermal facial landmark detector as well. In recent years, deep learning methods have been successfully applied to many areas such as image recognition due to their automatic learning capabilities, and have brought significant improvements in these areas. Therefore, this paper focuses on the thermal facial landmark detection based on deep learning methods.

Due to the expensive capture devices and limited dataset, there are only few works on thermal facial landmark detection. Active Appearance Models \cite{kopaczka2016robust} is a relatively traditional method initially used to model faces given facial landmark positions \cite{cootes1998active}. This method is a handcrafted model designed to compute the mean shape and appearance metric of a given dataset and use to it to predict the landmark of an unseen image. Like most handcrafted techniques, this method does not work well with high data variability.

A current trend towards thermal facial landmark detection is to apply deep learning models that are designed for detecting visible facial landmarks and adapted to the thermal image domain. Poster et al.\cite{poster2019examination} noted that the state-of-the-art visible facial landmark detection models work poorly with thermal images due to relatively low information contained in thermal images. This work shows that thermal images aligned with modern landmark detection algorithms often fail to achieve thermal-to-visible face verification results compared to manually aligned imagery. The models being tested in this study include Deep Alignment Network (DAN) \cite{kowalski2017deep}, Multi-task Convolutional Neural Network (MTCNN) \cite{zhang2016joint}, and Multi-class Patch-based fully convolutional neural network (PBC) \cite{poster2019examination}.

Another solution is to transform thermal images into visible images, and then to detect the landmarks from the transformed images using existing landmark detectors designed for visible images \cite{Chu2019}. The transformation is achieved by developing a CycleGAN model \cite{Zhu_2017_ICCV}. This method is shown to be unreliable, for example, CycleGAN mode collapse. This solution also offers room for inconsistency as it involves transforming a low information image into a higher information domain. To tackle this problem, an end-to-end landmark detection model based on two-stage training mechanism is proposed in \cite{Chu2019}. The first stage of this solution is to train a U-Net \cite{Ronneberger2015} to outline the face of a given image. A fully connected network is attached to the U-Net at the second stage to estimate facial landmark coordinates from the outline. An auxiliary output layer is used to detect the facial expression to enhance the training outcome. This solution however, are only able to detect front facing faces with no occlusion. Even self occlusion from slight face tilting to different direction results in landmark distortion.

Based on the aforementioned discussion, thermal facial landmark detection is still a challenging task. In this paper, we proposed a Deep Multi-Spectral Learning (DMSL) model for facial landmark detection. The proposed model is a stack of two independent sub-models, an auxiliary model for detecting face boundary, and a main model for detecting facial landmarks. DMSL is a unified model that can detect facial landmarks from both thermal and visible images, this leads to efficient and convenient solution for multi-spectral facial image analysis. Another core strength is that this model is robust for thermal facial landmark detection when a face is not front facing or when there is occlusion. Additionally, we annotated a thermal image dataset by a 68 points landmark configuration. This dataset is used for the experimentation of our proposed deep learning model based on U-Net to detect facial landmarks on thermal images. 

The rest of this paper is organized as follows. In Section 2, we present the proposed model architecture and pipeline. In section 3, we describe the dataset used for experimentation. The steps for pre-processing, and dataset annotation, are outlined in section 4. In section 5, we describe the metrics used to evaluate our model. In section 6, we explain the experiment process in detail as well as illustrate the performance and effectiveness of the model. Finally we conclude with a discussion in section 7.

\section{Multi-spectral facial landmark detector}
\subsection{Model intuition}
Compared to visible images, thermal images contain relatively low information and has a narrower pixel intensity range. When a network is being trained directly to detect facial landmarks, it is easy for the network to overfit and recognise area such as neck or large beard as facial landmarks. 

To address this problem we use an auxiliary model to highlight the face region within a given image and blackout other pixels first. This allows the main landmark detector to focus on the targeted region for more precise facial landmark detection.

\subsection{Model task formulation}
\begin{figure}[!htb]
\begin{center}
\includegraphics[width=0.49\linewidth]{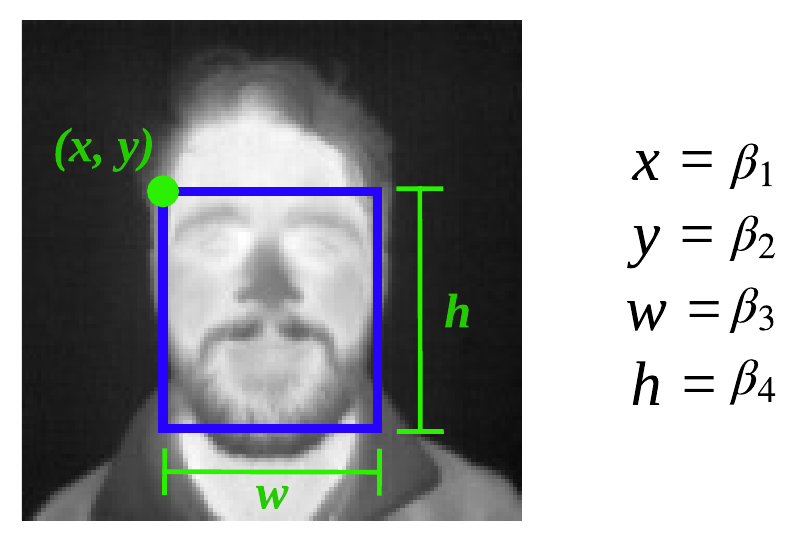}
\end{center}
   \caption{Array $B$ value depiction.}
\label{fig:beta_val}
\end{figure}

Let the auxiliary model of the DMSL that outputs face boundary be $M_b$. When given a face image $F$, the auxiliary outputs an array $B$ that contains four values $B=[\beta_1, \beta_2, \beta_3, \beta_4]$. These values represent the $(x, y)$ coordinate of the leftmost highest point, width $w$, and height $h$ of the boundary respectively. An example is depicted in figure \ref{fig:beta_val}.

\FloatBarrier

Boundary loss $L_{b}$ represented by equation \ref{eq:mse_bound} is used as the loss function for $M_b$:

\begin{equation}
    L_{b} = \frac{1}{V}\sum_{i=1}^{V} (\beta_{i} - \hat{\beta_{i}})^2
    \label{eq:mse_bound}
\end{equation}
where $V$ denotes the number of output values. In the case of face boundary detection, $V=4$. $\beta$ denotes the ground truth value and $\hat{\beta}$ denotes predicted value. This is the mean squared error between ground truth and predicted value.

\begin{figure}[!htb]
\begin{center}
\includegraphics[width=\linewidth]{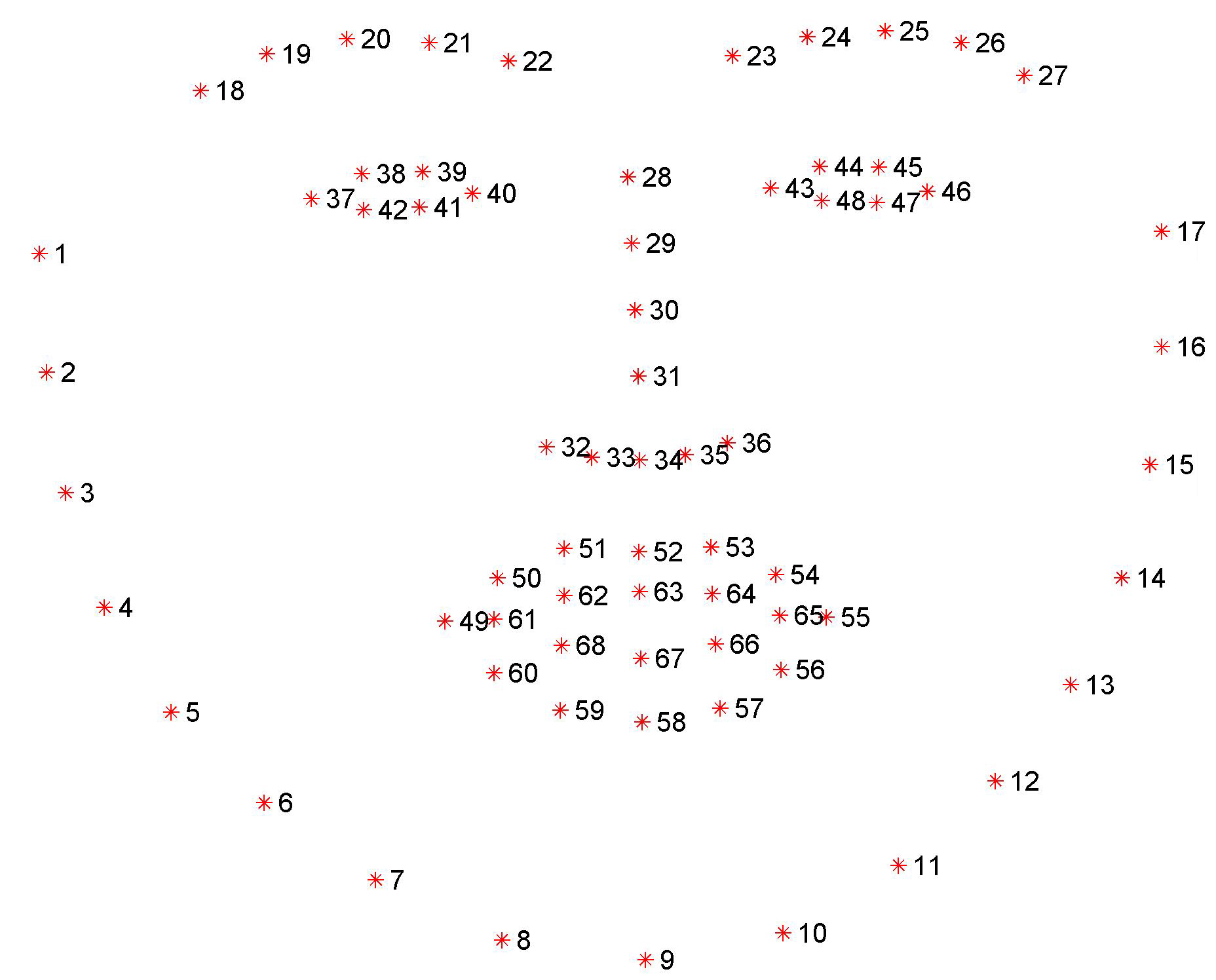}
\end{center}
  \caption{Illustration of $68$ point facial landmark configuration. Figure taken from \cite{sagonas2013300}.}
\label{fig:68point}
\end{figure}

The main model for facial landmark detection referred to as $M_l$ is trained to output an array $\Lambda$ that contains $P$ facial landmark coordinates when given a face images $F$. We used a well established 68 point facial landmark configuration from Multi-PIE \cite{gross2010multi} shown in figure \ref{fig:68point}. Each point is represented by an $(x, y)$ coordinate. $\Lambda$ is an array of $x$ and $y$ coordinates arranged in the form of $\Lambda=[x_1, y_1, x_2, y_2, x_3, y_3, ..., x_P, y_P]$, where $P = 68$.


\FloatBarrier

$M_l$ is trained using landmark loss $L_{l}$ implemented by \cite{Chu2019}. $L_{l}$ is defined as:

\begin{equation}
    L_{l} = \frac{1}{P}\sum_{i=1}^{P} ((x_{i} - \hat{x_{i}})^2 + (y_{i} - \hat{y_{i}})^2)
    \label{eq:l1}
\end{equation}
where $P$ denotes the number of landmark points. This calculates with the squared difference between ground truth coordinates $\Lambda$, and predicted coordinates $\hat{\Lambda}$. 







\subsection{Model architecture}
We use a modified version of U-Net developed by \cite{Ronneberger2015}. U-Net is originally used for biomedical image segmentation to tackle the issue of limited availability of data within the domain. This solves our problem of having only small datasets. Another reason to use U-Net is because our task is similar to image segmentation. In biomedical image segmentation, the goal is for the model to outline the target area required by user. This fits our purpose because we need to outline the prominent landmark features on the face and then to find the corresponding coordinates. 

\begin{figure}[!htb]
\begin{center}
\includegraphics[width=\linewidth]{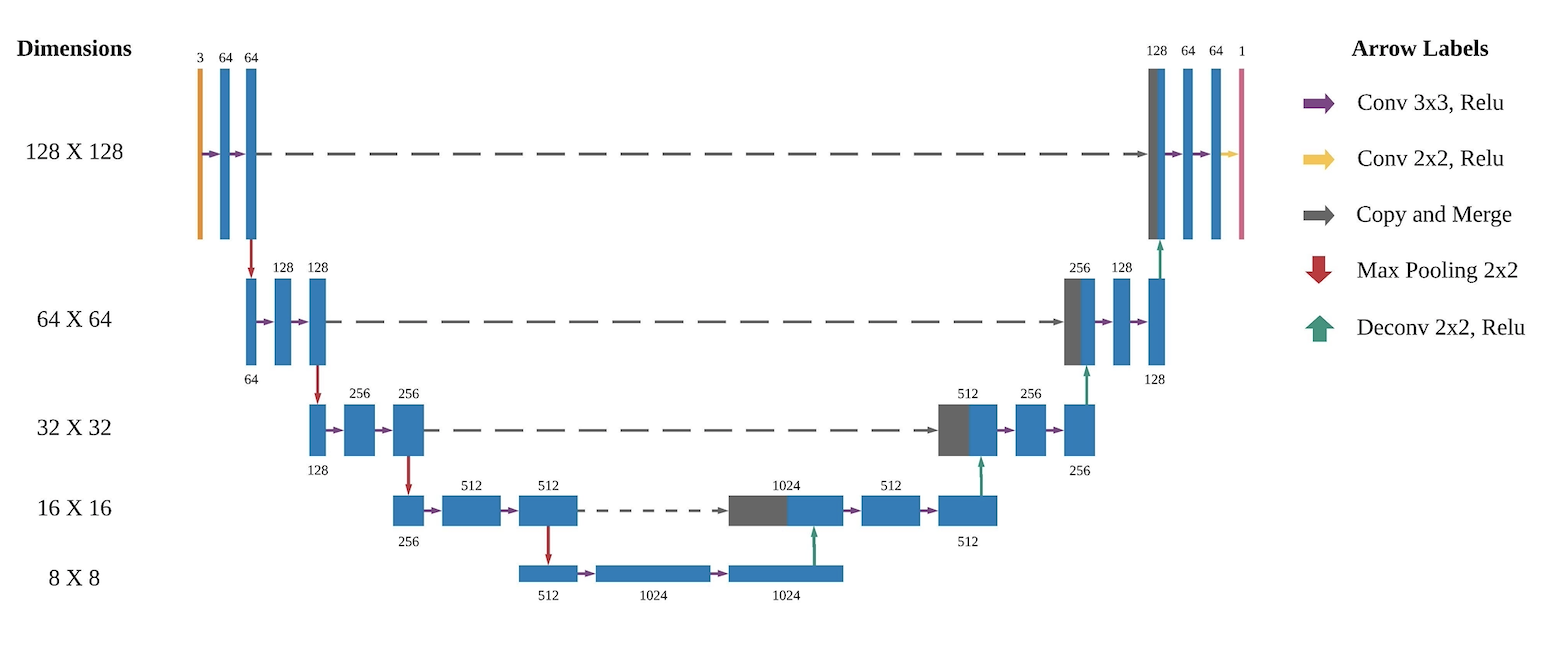}
\end{center}
   \caption{The configured U-Net structure.}
\label{fig:unet}
\end{figure}
\FloatBarrier
U-Net is a neural network that is only composed of convolutional layers. There are two major paths in this architecture, down-sampling and up-sampling. Figure \ref{fig:unet} shows our configuration of U-Net. The left side of the U shape structure is the down-sampling path, which is similar to typical convolutional layers in convolutional neural network (CNN). One down-sampling step involves two $3\times3$, stride 1 convolution and a $2\times2$ max pooling with stride $2$. An up-sampling step starts with a $2\times2$ deconvolution with stride $2$, and followed by two $3\times3$, stride 1 convolution. At every up-sampling step, features from the down-sampling path with the same dimension are concatenated to the up-sampled feature. This is represented by the gray arrow in figure \ref{fig:unet}. Finally, we have a $1\times1$, stride 1 convolution operation to create an output with the same dimension as input but with depth set to one. With this configuration, the output dimension should be the same as the input, which in our case is $128\times128$. The activation function used throughout is ReLU.

\begin{figure}[!htb]
\begin{center}
\includegraphics[width=\linewidth]{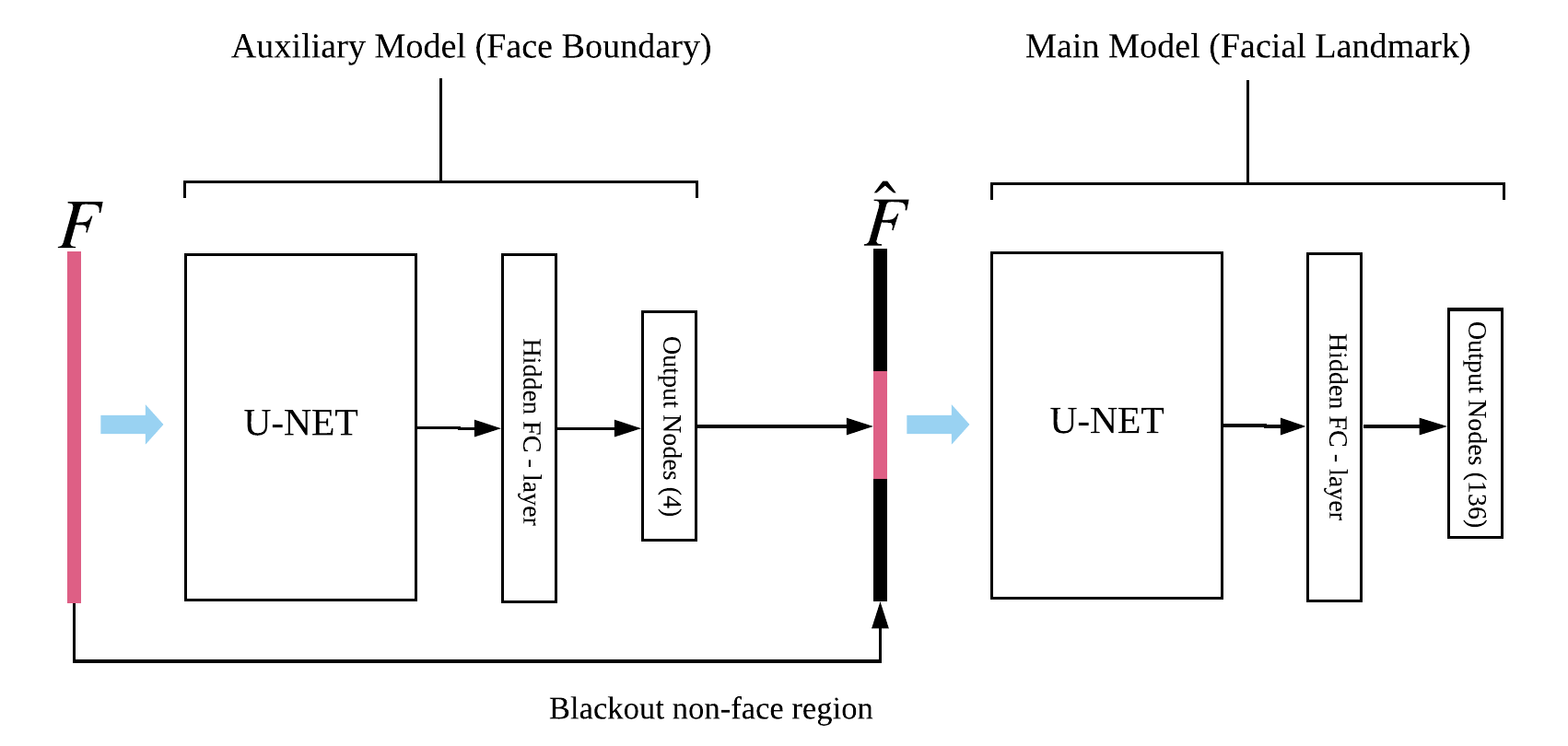}
\end{center}
   \caption{Illustration of our proposed pipeline.}
\label{fig:pipeline}
\end{figure}
\FloatBarrier

Figure \ref{fig:pipeline} shows the pipeline of DMSL. $F$ represents an input image and $\hat{F}$ represents the image with non-face region blacked out using the output from the auxiliary model. From figure \ref{fig:pipeline}, Auxiliary model illustrates $M_b$. The output dimension of U-Net in $M_b$ is flatted to $16384\times1$. It is then connected to a fully connected network with a hidden layer with $1024$ nodes and an output layer with $4$ nodes for predicting array $B$. $M_l$ is represented by the main model in figure \ref{fig:pipeline}. The difference between the two models is that the output layer in $M_l$ has $136$ nodes for the prediction of array $\Lambda$. The sigmoid function is used for all fully connected layers.

\section{Dataset}
Eurecom's VIS-TH visible and thermal paired face database (VIS-TH) is used in this study \cite{Mallat18}. The dataset is composed of $50$ subjects with varying ethnicity, sex, and age. The camera is set to capture two images simultaneously, one capturing EMR from the thermal spectrum and another capturing visible EMR. 

\begin{figure}[!htb]
\begin{center}
\includegraphics[width=\linewidth]{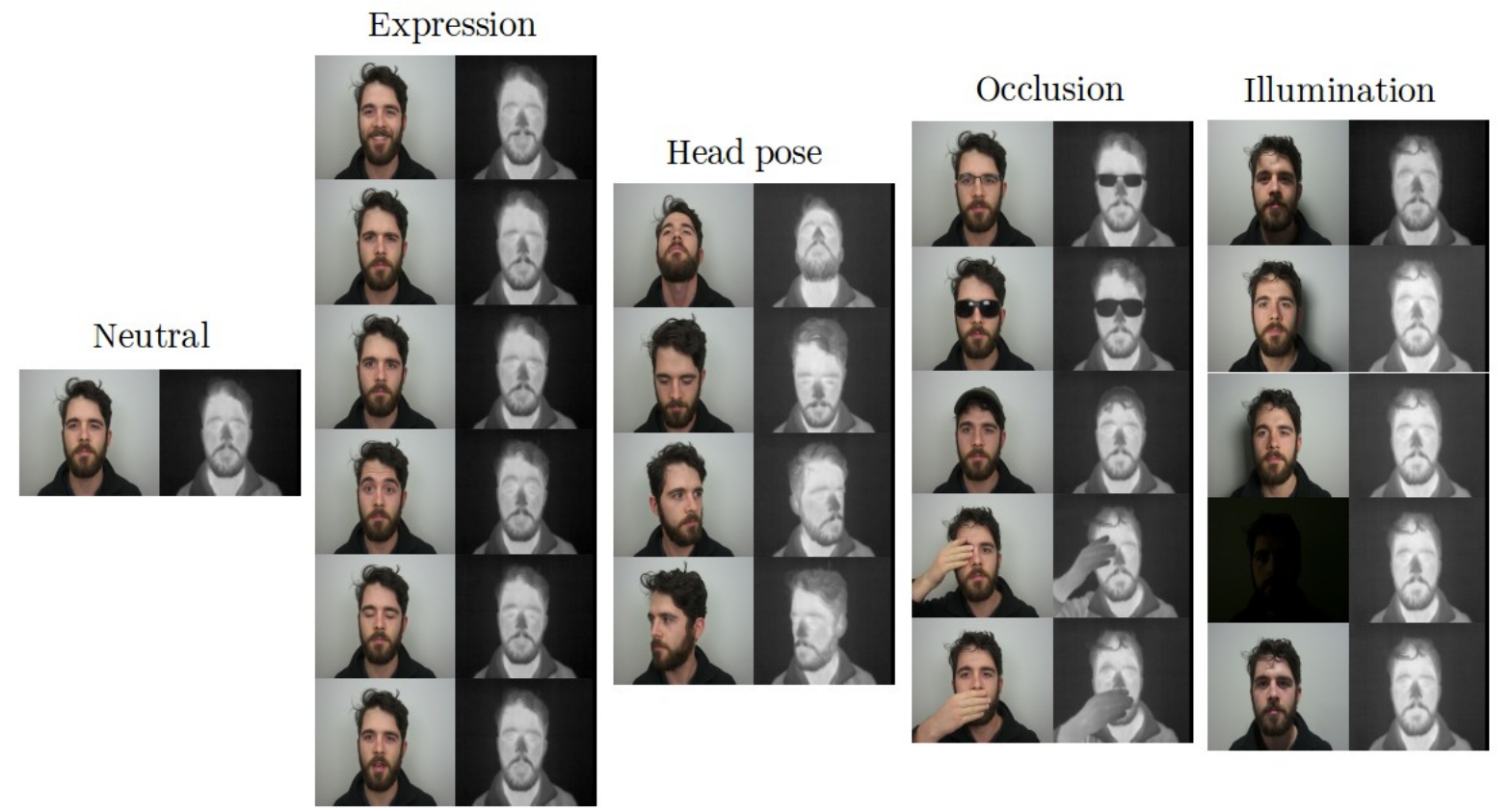}
\end{center}
   \caption{Example of visible and thermal image pair with all variations. Figure taken from \cite{Mallat18}}
\label{fig:variation}
\end{figure}
\FloatBarrier

The image pair for all variations is shown in figure \ref{fig:variation}, and all variations are listed in table \ref{tab:variations}. Each subject has $21$ visible (VIS) images and $21$ thermal (TH) images. This results in a total of $2100$ images included in the dataset. It is worthy to highlight that VIS-TH dataset fits our settings due to the pixel-to-pixel registered images for both VIS and TH face. This makes it convenient for annotation purposes.

\begin{table}[!htb]
\centering
\begin{tabular}{lll}
\hline
Variation       & Type       & Acronym \\ \hline
Neutral         & Expression & NN      \\
Happy           & Expression & EH      \\
Angry           & Expression & EA      \\
Sad             & Expression & ES      \\
Surprised       & Expression & ESp     \\
Eyes closed     & Action     & AEC     \\
Open mouth      & Action     & AOM     \\
Look up         & Pose       & PU      \\
Look down       & Pose       & PD      \\
Look left       & Pose       & PL      \\
Look right      & Pose       & PR      \\
Optical glasses & Occlusion  & OOG     \\
Sunglasses      & Occlusion  & OSG     \\
Hat             & Occlusion  & OH      \\
Hand on mouth   & Occlusion  & OHM     \\
Hand on eye     & Occlusion  & OHE     \\
Light up        & Light      & LLU     \\
Light right     & Light      & LLR     \\
Light left      & Light      & LLL     \\
Dark            & Light      & LD      \\
Room light      & Light      & LR      \\ \hline
\end{tabular}
\caption{List of variations.}
\label{tab:variations}
\end{table}
\FloatBarrier

\section{Pre-prossessing, Augmentation, and Annotation}
\begin{figure}[htb!]
\begin{center}
  \imgStub{0.14}{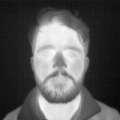}{ (a) Original image } 
  \imgStub{0.14}{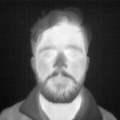}{(b) Mirrored image }
   \caption{Example of original and mirrored image.\label{fig:augmentation}}
\end{center}
\end{figure}

All images are cropped into a $1:1$ ratio image without cutting part of the subject face. This is because we need to match the input dimension of the U-Net convolution layer $(120 \times 12)$. To mitigate the issue of having a small dataset, we generate a mirrored version for each image within the dataset. The result is shown in figure \ref{fig:augmentation}. This increases the final number of images to $4200$. There is a particular variation where an image is taken under all lights off condition. The resulting VIS image is just a black image so we can omit this variation in the experiments, making the total usable images count $4100$. 

\FloatBarrier

\begin{figure}[!htb]
\begin{center}
\includegraphics[width=\linewidth]{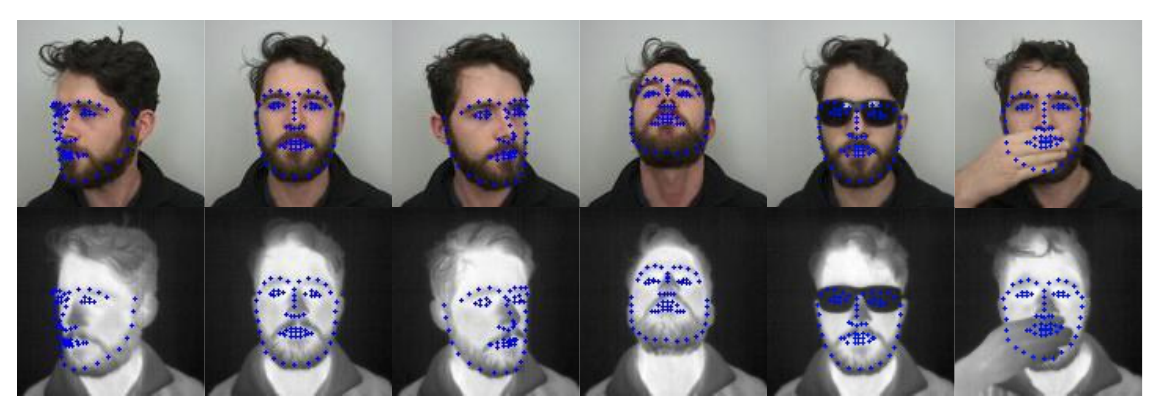}
\end{center}
   \caption{Example of annotated images.}
\label{fig:annotated}
\end{figure}

We use Dlib-ml \cite{King2009} to annotate facial landmark on all VIS images. Taking advantage of this dataset having both TH and VIS image being captured simultaneously, the annotated data is subsequently superimposed on their TH image counterpart. This gives us ground truth array $\Lambda$. Figure \ref{fig:annotated} shows the landmark plot for multiple images. The first row shows the VIS images being used for detection, while the second row represents plots of the superimposed coordinates on TH image. This is used as ground truth array $B$.

\FloatBarrier

In many cases, there are some misalignment of landmark points by Dlib. To counter this, we manually calibrate inaccurate points to fit the face within an image. From the adjusted landmark coordinates, we find the leftmost highest point, rightmost point and, the lowest point from it. From this, we can calculate the width and height of face. These value along with the two coordinate value from the leftmost highest point is used to represent the boundary of face.
\section{Evaluation Metrics}

The performance of DMSL is evaluated by measuring normalized mean error (NME) used by \cite{sagonas2013300}. It normalises the distance between ground truth and predicted value with the distance of both eyes. This is represented by equation \ref{eq:nme}

\begin{equation}
    NME = \frac{1}{N}\sum_{i=1}^{N}\frac{\left \| \Lambda_i - \hat{\Lambda_i} \right \|}{P \times D_i}
    \label{eq:nme}
\end{equation}

In equation \ref{eq:nme}, $\left \| \Lambda - \hat{\Lambda} \right \|$ represents the L2 norm between ground truth, $\Lambda$ and predicted value, $\hat{\Lambda}$. $D_i$ represents the inter-ocular distance for that particular face. This is calculated by the euclidean distance between the outer corner of two eyes. For our configuration, the right eye is represented by point $37$ while the left eye is marked with point $46$ as illustrated in figure \ref{fig:68point}. $P$ denotes the number of facial landmark points. Finally, $N$ represents the number of sample images used.


\section{Experiment and results}
\subsection{Training and testing protocol}

\begin{figure}[!htb]
\begin{center}
\includegraphics[width=.6\linewidth]{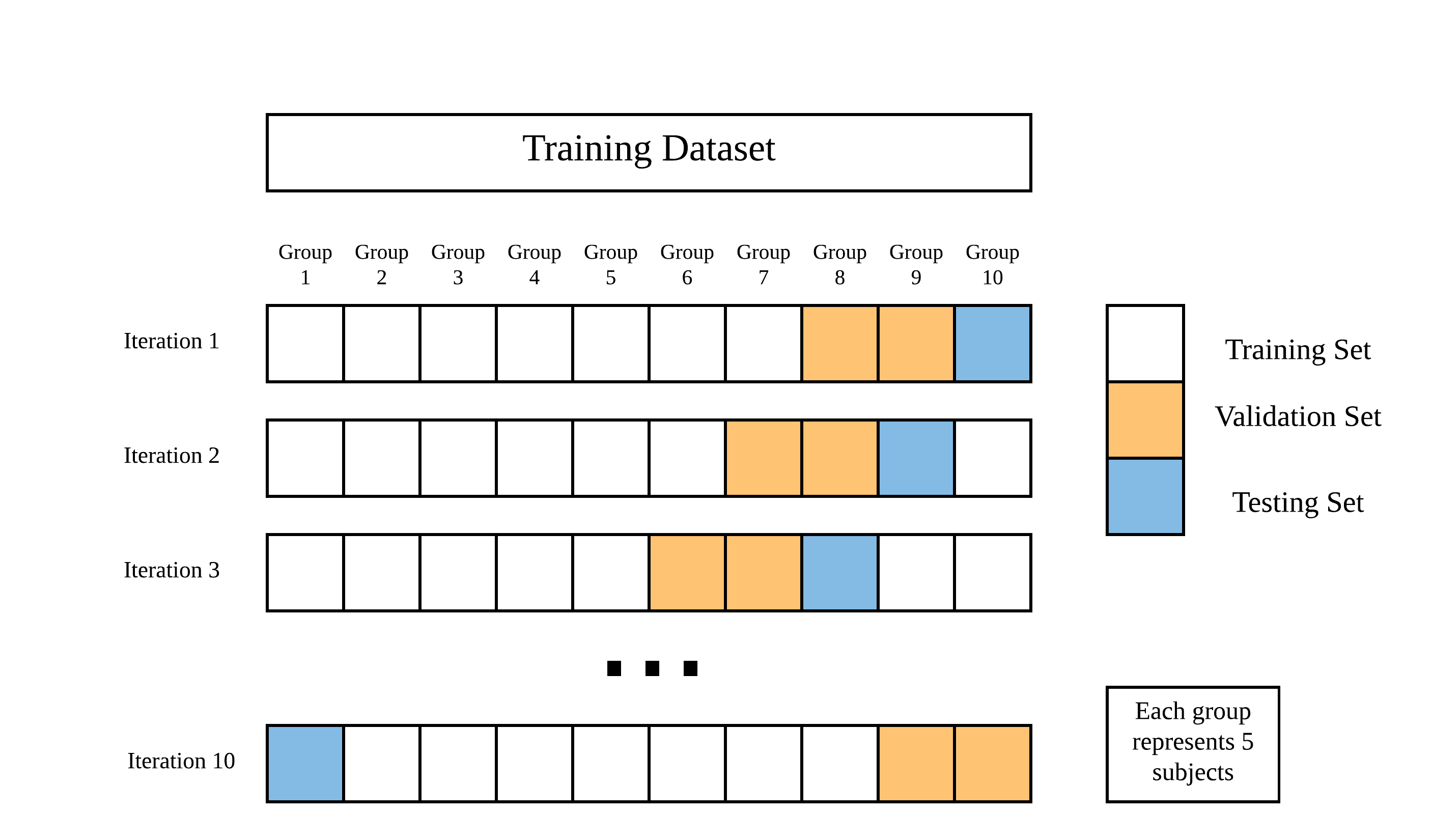}
\end{center}
  \caption{10-fold cross validation process.}
\label{fig:fold}
\end{figure}

We train and test the model using a 10-fold cross validation. The $50$ subjects are divided into $10$ groups by their ID number. Each iteration we use $1$ group for testing, $2$ for validating, and the rest for training. This process is illustrated in figure \ref{fig:fold}.

\FloatBarrier

The experiments are conducted on a Linux server equipped with one NVIDIA Titan Xp graphics processing unit (GPU). The deep learning models are constructed and tested with TensorFlow 2.0.0-rc0 using Python 3.7.

\subsection{Training}
To produce an optimised DMSL model, Both $M_b$ and $M_l$ have to be trained individually using a two stage training procedure similar to the training process in \cite{Chu2019}. This is to maximise the advantage of using a U-Net as feature extractor. In essence, we have two individual models and both require a two stage training procedure.

\subsubsection{Face boundary detection model \boldmath$M_b$}
\begin{figure}[htb!]
\begin{center}
  \imgStub{0.14}{Images/TH_001_1_01_NN.jpg}{ (a) TH image } 
  \imgStub{0.14}{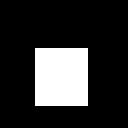}{(b) Ground truth }
  \imgStub{0.14}{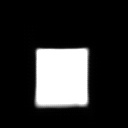}{(c) Predicted}
   \caption{Example of face boundary mask.\label{fig:boundarymask}}
\end{center}
\end{figure}

The first stage is to train a U-Net to outline the face boundary in the original image. A boundary mask shown in figure \ref{fig:boundarymask}(b) is generated from our face boundary ground truth annotation. 

The goal is to train the U-Net to get an output similar to the boundary mask shown in figure \ref{fig:boundarymask}(b). Figure \ref{fig:boundarymask}(c) shows the output by this U-Net. This is achieved by minimising U-Net loss defined in equation \ref{eq:l_unet}. 

\begin{equation}
    L_u = -\frac{1}{N}\sum_{i=1}^{N} m \log \hat{m} + (1-m) \log (1-\hat{m})
    \label{eq:l_unet}
\end{equation}
where, $m$, and $\hat{m}$ represents ground truth mask and predicted mask respectively with $N$ represents the number of used sample images. This computes the softmax cross-entropy loss between the two masks.

The second stage is for training the attached fully connected network to output the $B$ array. The parameters of the convolutional U-Net are frozen and the fully connected network is trained by minimising boundary loss represented by equation \ref{eq:mse_bound}. 

\begin{figure}[htb!]
\begin{center}
  \imgStub{0.11}{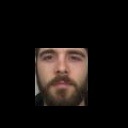}{ (a) Ground truth VIS} 
  \imgStub{0.11}{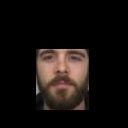}{(b) Predicted VIS}
  \imgStub{0.11}{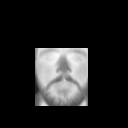}{(c) Ground truth TH}
  \imgStub{0.11}{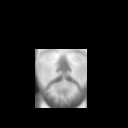}{(d) Predicted TH}
   \caption{Example of non-face region blacked out using ground truth value and $M_b$ generated value.\label{fig:cropped}}
\end{center}
\end{figure}
\FloatBarrier

Figure \ref{fig:cropped} shows a comparison between images cropped using values predicted by the auxiliary model against values from ground truth. These cropped images will be the input of the main model that detects facial landmark. 

\subsubsection{Facial landmark detection model \boldmath$M_l$}
\begin{figure}[htb!]
\begin{center}
  \imgStub{0.15}{Images/TH_001_1_01_NN.jpg}{ (a) TH image } 
  \imgStub{0.15}{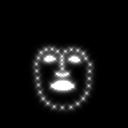}{(b) Ground truth }
  \imgStub{0.15}{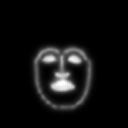}{(c) Predicted}
   \caption{Example of facial landmark mask.\label{fig:landmarkmask}}
\end{center}
\end{figure}

The main model uses the same two stage training procedure. The labels for first stage are mask generated from the landmark coordinates with algorithm \ref{alg:maskgen}. Figure \ref{fig:landmarkmask}(b) is an example of the mask generated.

\FloatBarrier

Algorithm \ref{alg:maskgen} shows how the mask is generated, it is based on the method used by \cite{Chu2019}. The idea behind this mask is to highlight all the $68$ facial landmark points. The surrounding area are set to gradually dim creating a glowing effect. This helps the model to outline the facial landmarks. 

\begin{algorithm}[!htb]
\SetAlgoLined
\SetKwData{Left}{left}\SetKwData{This}{this}\SetKwData{Up}{up}
\SetKwFunction{Union}{Union}\SetKwFunction{FindCompress}{FindCompress}
\SetKwInOut{Input}{input}\SetKwInOut{Output}{output}
\Input{Coordinates, $Dimension_x$, $Dimension_y$}
\Output{Mask}
\BlankLine
Initialize 2D array (Mask) according to $Dimension_x$, and $Dimension_y$\;
    \While{$i \leq length(Coordinates)$}{
        $x \gets Coordinates[i] \times Dimension_x$\;
        $y \gets Coordinates[i+1] \times Dimension_y$\;
        \For{$j=1, j \leq X-Dimension$}{
            \For{$k=1, j \leq Y-Dimension$}{
                \If{$Mask[j][k] < 255$}{
                    $Mask[j][k] += 0.5^{\max{(\left | x - j \right |, \left | y - k \right |)}}\times255$\;
                }
            }
        }
 }
 \caption{Generate Landmark Mask}
 \label{alg:maskgen}
\end{algorithm}
\FloatBarrier

The goal in stage one is to train a U-Net to output an image similar to Figure \ref{fig:landmarkmask}(b). The training is done by minimising U-Net loss. Figure \ref{fig:landmarkmask}(c) shows an example of the U-Net output.

Stage two is for training the attached fully connected network of the main model to output the $\Lambda$ array. The convolutional U-Net parameters are frozen, while the fully connected network are trained by minimising the loss represented by equation \ref{eq:l1}.

\subsection{Performance of Landmark Detection}
\begin{figure}[!htb]
\begin{center}
\includegraphics[width=0.75\linewidth]{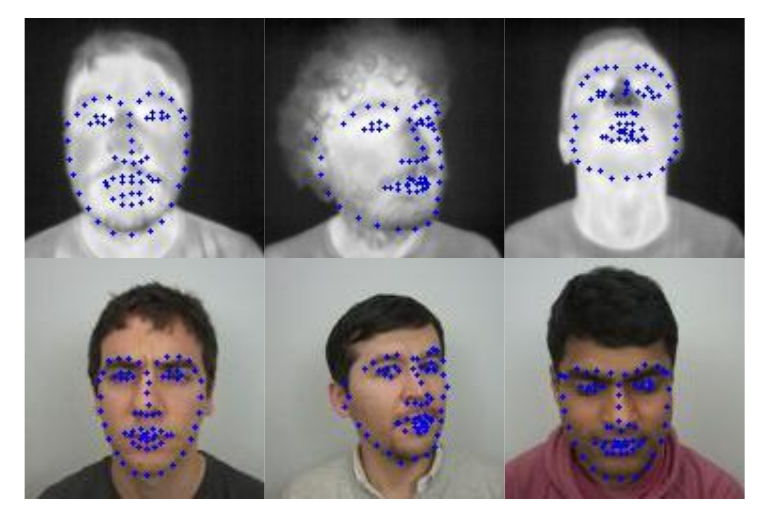}
\end{center}
   \caption{Sample plots of output from our model with different face pose.}
\label{fig:output_pose}
\end{figure}

Figure \ref{fig:output_pose} shows the plotted facial landmark output by our model. The model is able to detect facial landmarks with samples facing different directions. 

\FloatBarrier

\begin{figure}[!htb]
\begin{center}
\includegraphics[width=.9\linewidth]{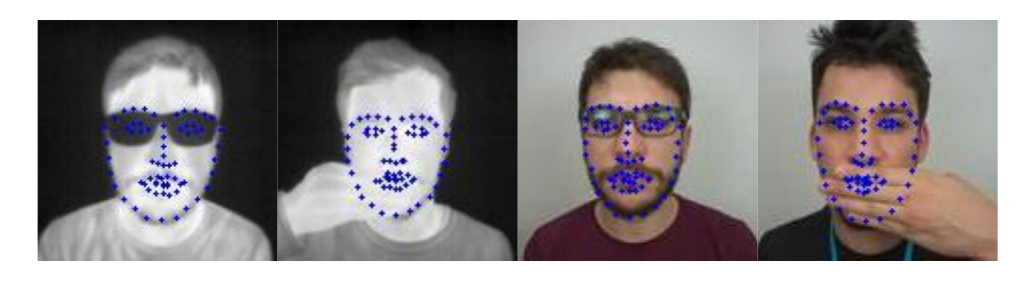}
\end{center}
   \caption{Sample plots of output from our model with different face occlusions.}
\label{fig:output_occ}
\end{figure}

The model also works fairly well when certain areas such as eyes or mouth are occluded. This is shown in figure \ref{fig:output_occ}.

\FloatBarrier

We evaluate the performance of our model against Dlib-ml \cite{King2009}, the optimal AAM method examined in \cite{kopaczka2016robust}, and a two stage fully connected multi-task U-Net \cite{Chu2019}. The output of both models is evaluated using our manually annotated ground truth. The comparison is shown in table \ref{tab:result}. Our model is able to achieve a better result for TH images and a competitive result for VIS images. For comparison purposes, we used the samples with landmark coordinates that we adjusted (for more accurate ground truth) to compute error values. We also excluded the samples that Dlib-ml and AAM models are unable to detect. By doing so, we ensures a fair comparison of flaws. These samples consist of $711$ TH images and $139$ VIS images including the mirrored version.

\begin{table}[!htb]
\begin{center}
\begin{tabular}{|l|c|c|}
\hline
 & TH & VIS \\ \hline
AAM \cite{kopaczka2016robust} & 0.1311 & 0.1434 \\ 
Dlib \cite{King2009} & 0.0293 & 0.0581 \\ 
Chu et al. \cite{Chu2019} & 0.0222 & 0.0556 \\
Ours & 0.0210 & 0.0544 \\ \hline
\end{tabular}
\end{center}
\caption{Model performance comparison in terms of NME}
\label{tab:result}
\end{table}
\FloatBarrier

\begin{figure}[!htb]
\begin{center}
\includegraphics[width=.8\linewidth]{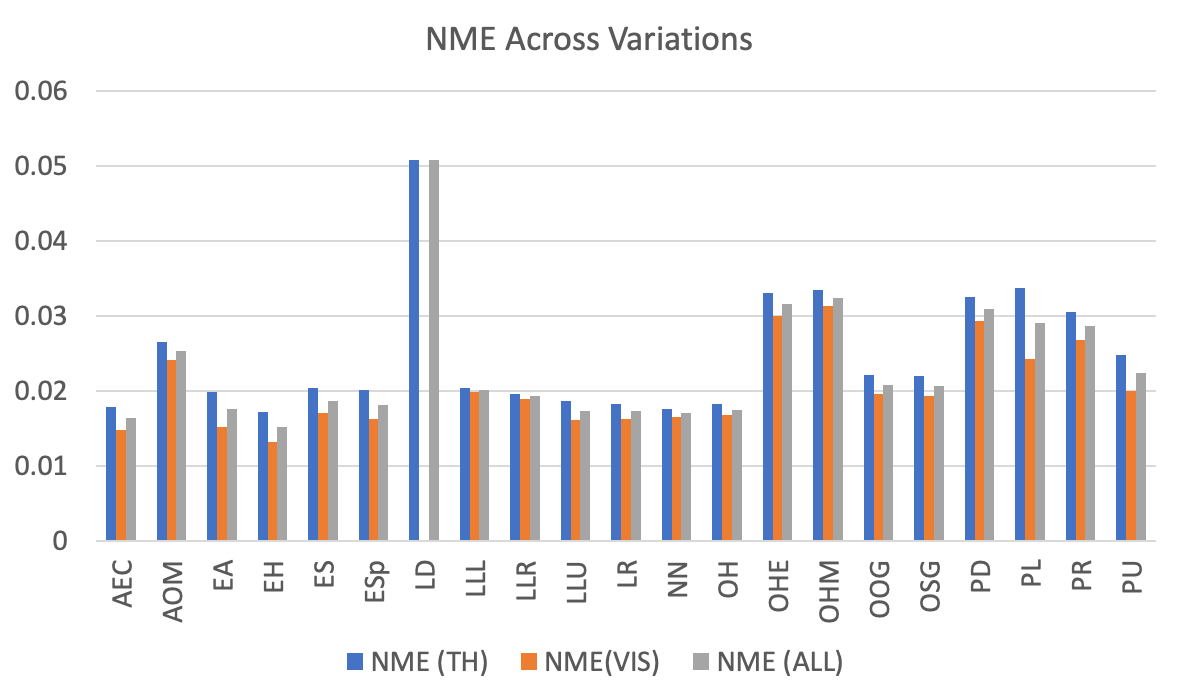}
\end{center}
   \caption{Normalised mean error comparison between variations.}
\label{fig:nme}
\end{figure}

Figure \ref{fig:nme} illustrates the performance of our DMSL model in detail. This shows the average error value for each variation with all available images included. The variation labels are listed in table \ref{tab:variations}.

\FloatBarrier



\section{Discussion and conclusion}
An interesting observation from the experiments is that by training model with images from both spectrum yields better results for thermal landmark detection. This is illustrated in figure \ref{fig:nme} where the class $LD$ have relatively poor performance when compared to other variations. This class is trained with a lack of a usable VIS image.

\begin{figure}[!htb]
\begin{center}
\includegraphics[width=.4\linewidth]{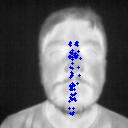}
\end{center}
   \caption{Ill-shaped facial landmark output.}
\label{fig:compressed}
\end{figure}

Experimental results have shown that our DMSL model can outperform existing landmark detection methods. However, we still have a small amount of cases where the model outputs ill-shaped or compressed landmark coordinates. An example of this output is illustrated in figure \ref{fig:compressed}. This could be potentially mitigated in the future by experimenting with more loss functions and adjusting the fully connected network.

\FloatBarrier

The main contribution of this study is that we have developed a DMSL model for facial landmark detection This model can detect facial landmarks in TH and VIS images regardless of which direction the target is facing. In addition, the model can still detect landmark on faces that are partially occluded by objects such as sunglasses, hand, and hat. Both of this capability are not present in existing state-of-the-art thermal facial landmark systems. Another contribution of this study is to provide a facial landmark annotation for a TH image dataset \cite{Mallat18}.

This study can be further applied for an array of future research involving multi-spectral or thermal imagery. This includes but not limited to facial recognition, thermal-to-visible face generation, and real-time surveillance. Since we are using information collected from heat generated by a human body, this research can be potentially integrated with facial biometrics tasks with body temperature detection. This could help with public health investigation for disease detection. 


{\small
\bibliographystyle{ieee}
\bibliography{ThermalLandmark}
}

\end{document}